\documentclass[conference]{IEEEtran}
\IEEEsettopmargin{t}{72pt}
\IEEEoverridecommandlockouts
\usepackage{cite}
\usepackage{amsmath,amssymb,amsfonts}
\usepackage{algorithmic}
\usepackage{graphicx}
\usepackage{textcomp}
\usepackage{xcolor}
\usepackage{hyperref}
\usepackage{subcaption} 
\usepackage{booktabs}
\def\BibTeX{{\rm B\kern-.05em{\sc i\kern-.025em b}\kern-.08em
    T\kern-.1667em\lower.7ex\hbox{E}\kern-.125emX}}

\begin{document}

\title{\LARGE \bf
Enhanced View Planning for Robotic Harvesting: Tackling Occlusions with Imitation Learning
}

\author{
Lun Li and Hamidreza Kasaei
\thanks{
Lun Li and Hamidreza Kasaei both are with the Department of Artificial Intelligence, Bernoulli Institute, Faculty of Science and Enginerring, University of Groningen, The Netherlands
}
\thanks{
Emails: \{alan.li, hamidreza.kasaei\}@rug.nl
}
}

\maketitle

\begin{abstract}

In agricultural automation, inherent occlusion presents a major challenge for robotic harvesting. We propose a novel imitation learning-based viewpoint planning approach to actively adjust camera viewpoint and capture unobstructed images of the target crop. Traditional viewpoint planners and existing learning-based methods, depend on manually designed evaluation metrics or reward functions, often struggle to generalize to complex, unseen scenarios. Our method employs the Action Chunking with Transformer (ACT) algorithm to learn effective camera motion policies from expert demonstrations. This enables continuous six-degree-of-freedom (6-DoF) viewpoint adjustments that are smoother, more precise and reveal occluded targets. Extensive experiments in both simulated and real-world environments, featuring agricultural scenarios and a 6-DoF robot arm equipped with an RGB-D camera, demonstrate our method's superior success rate and efficiency, especially in complex occlusion conditions, as well as its ability to generalize across different crops without reprogramming. This study advances robotic harvesting by providing a practical “learn from demonstration” (LfD) solution to occlusion challenges, ultimately enhancing autonomous harvesting performance and productivity.

\end{abstract}

\section{Introduction}

Occlusion, particularly when the target crop is obscured by surrounding crops or foliage, remains one of the most significant challenges for automated harvesting in agricultural robotics \cite{gong2022robotic}. This issue originates from the natural growth characteristics of plants, which have evolved to shield fruits behind leaves to protect them from direct sunlight and harsh weather, preventing scalding or damage. However, this protective trait also leads to occlusion of both fruits and stems, complicating automated harvesting. Gongal et al. \cite{gongal2015sensors} reported that occlusion can result in a fruit detection failure rate of up to 30\% in orchard environments, which reduces harvesting efficiency and increases the risk of crop damage and yield loss. Thus, addressing occlusion is essential for improving the performance of autonomous harvesting systems.

Humans instinctively adjust their posture and viewpoint to gather sufficient information when faced with occlusions. This concept of active vision was first introduced by Aloimonos et al. \cite{aloimonos1988active}, who demonstrated that many computer vision problems could be simplified by actively changing the observer’s position relative to the environment. In robotics, this is referred to as view planning or the next-best-view (NBV) problem, wherein deliberate decisions about camera viewpoints can help mitigate occlusions in robotic harvesting \cite{magalhaes2022active}.

\begin{figure}[htpb]

\centerline{
\includegraphics[width=0.239\textwidth]{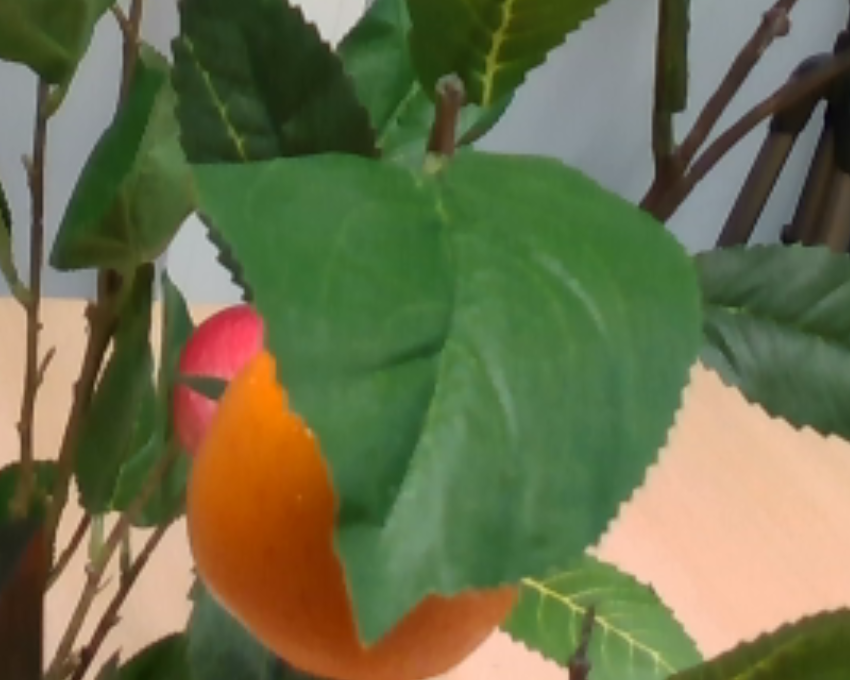}
\includegraphics[width=0.239\textwidth]{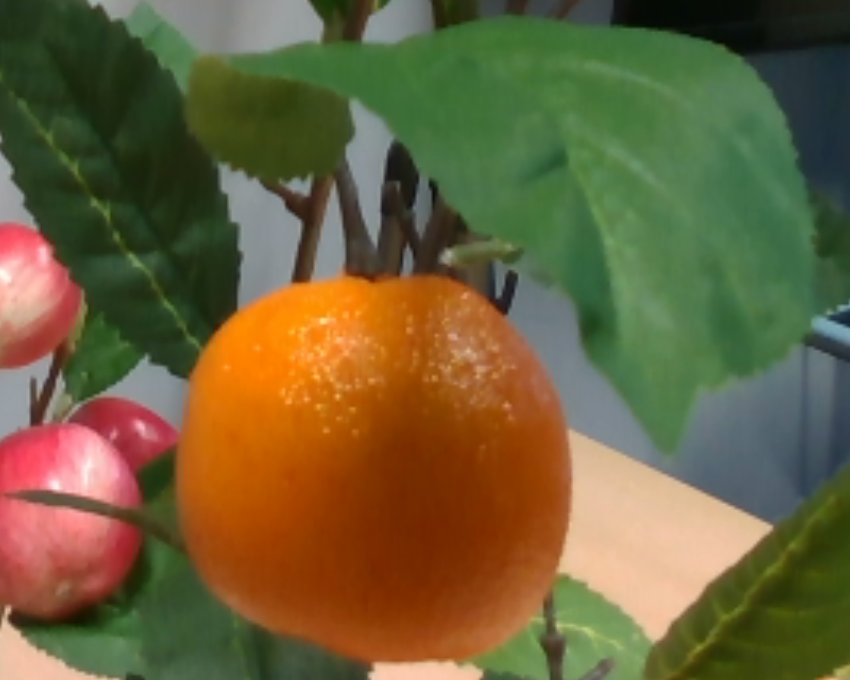}
}
\vspace{3pt}

\centerline{
\includegraphics[width=0.116\textwidth]{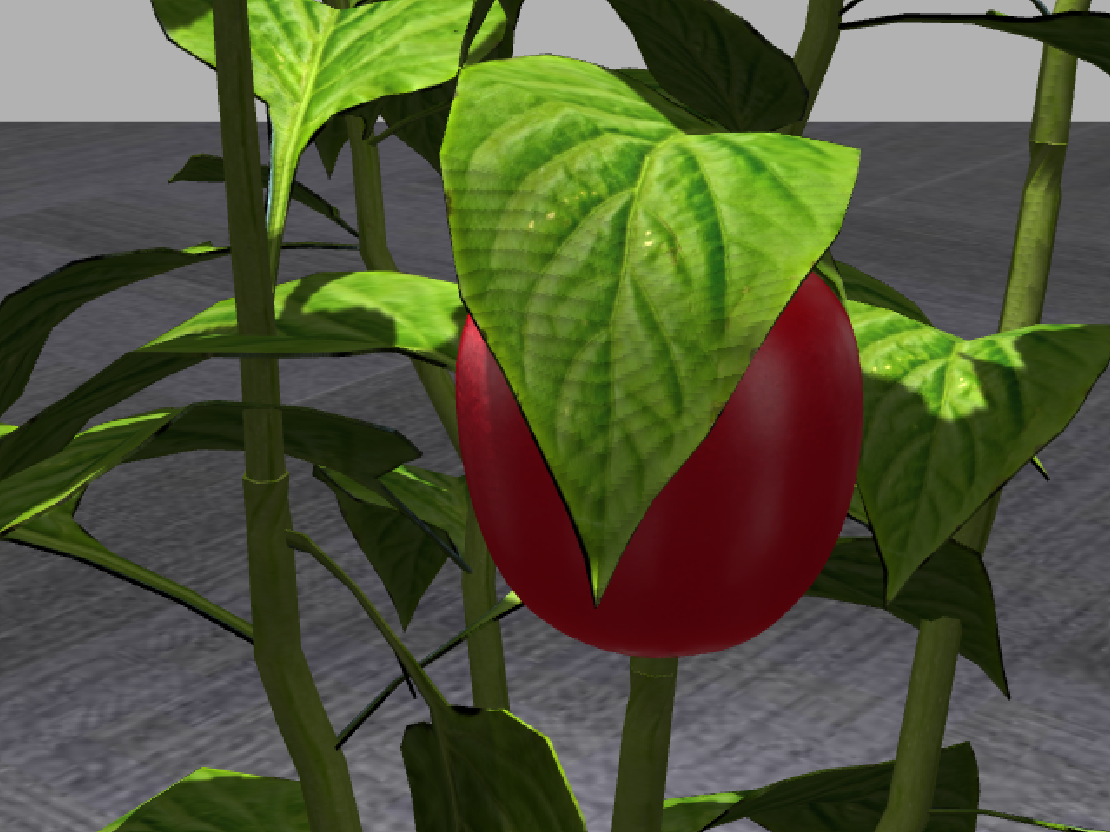}
\includegraphics[width=0.116\textwidth]{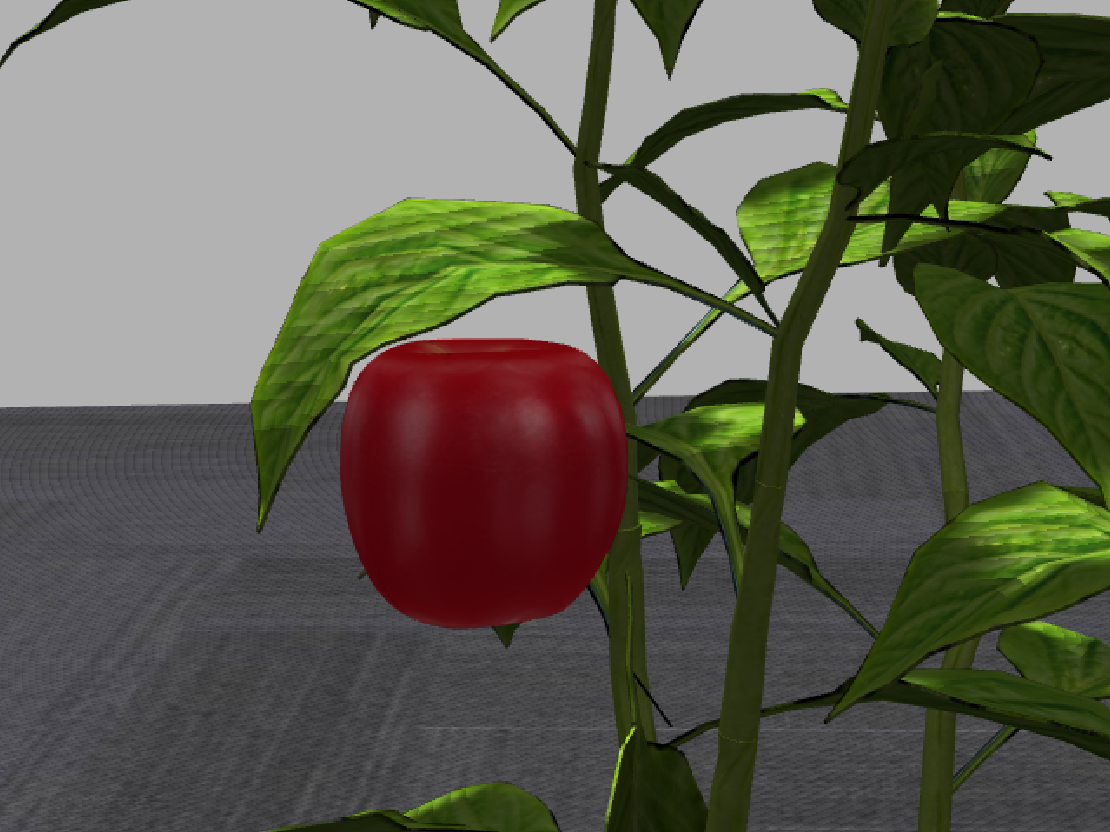}
\includegraphics[width=0.116\textwidth]{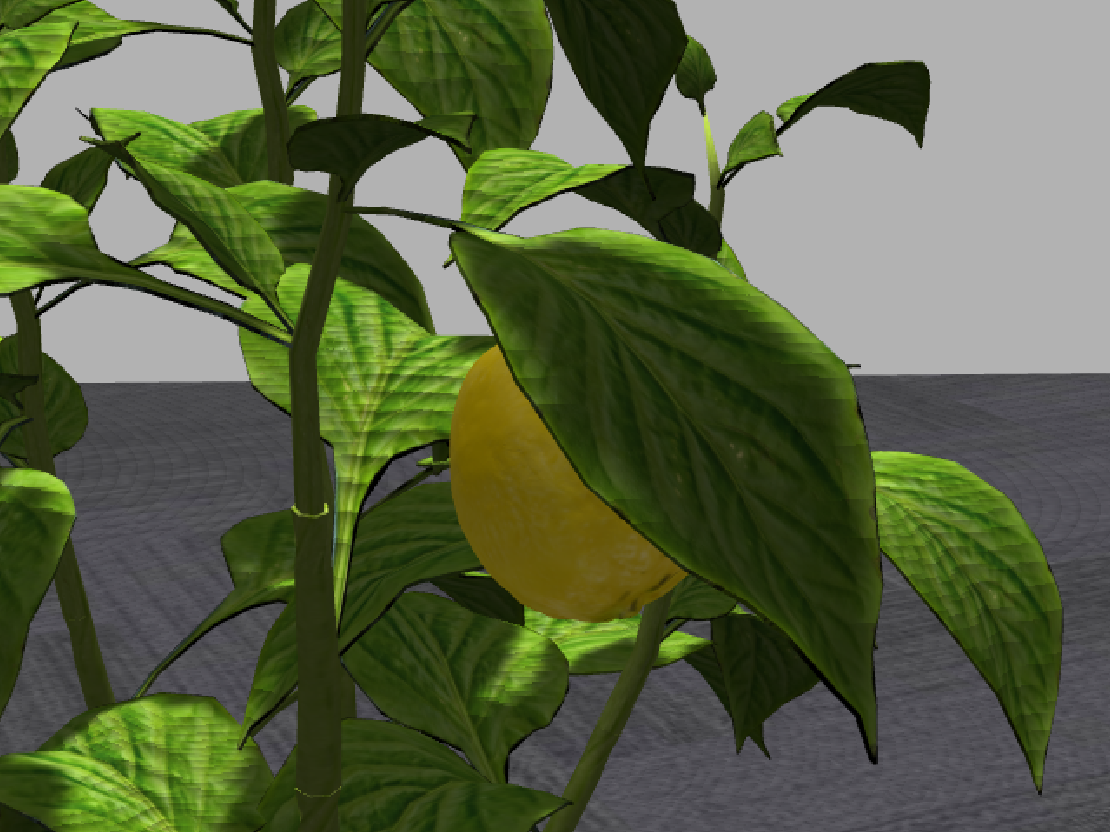}
\includegraphics[width=0.116\textwidth]{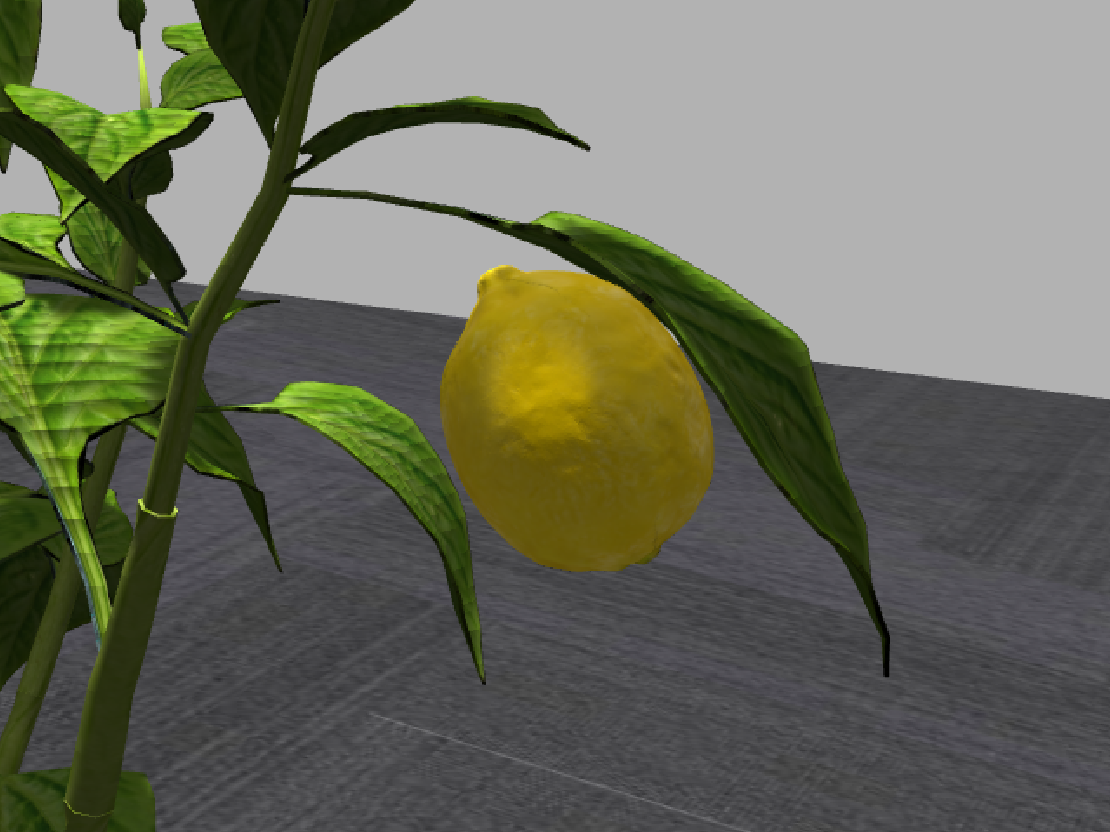}

}

\vspace{3pt}
\centerline{
\includegraphics[width=0.116\textwidth]{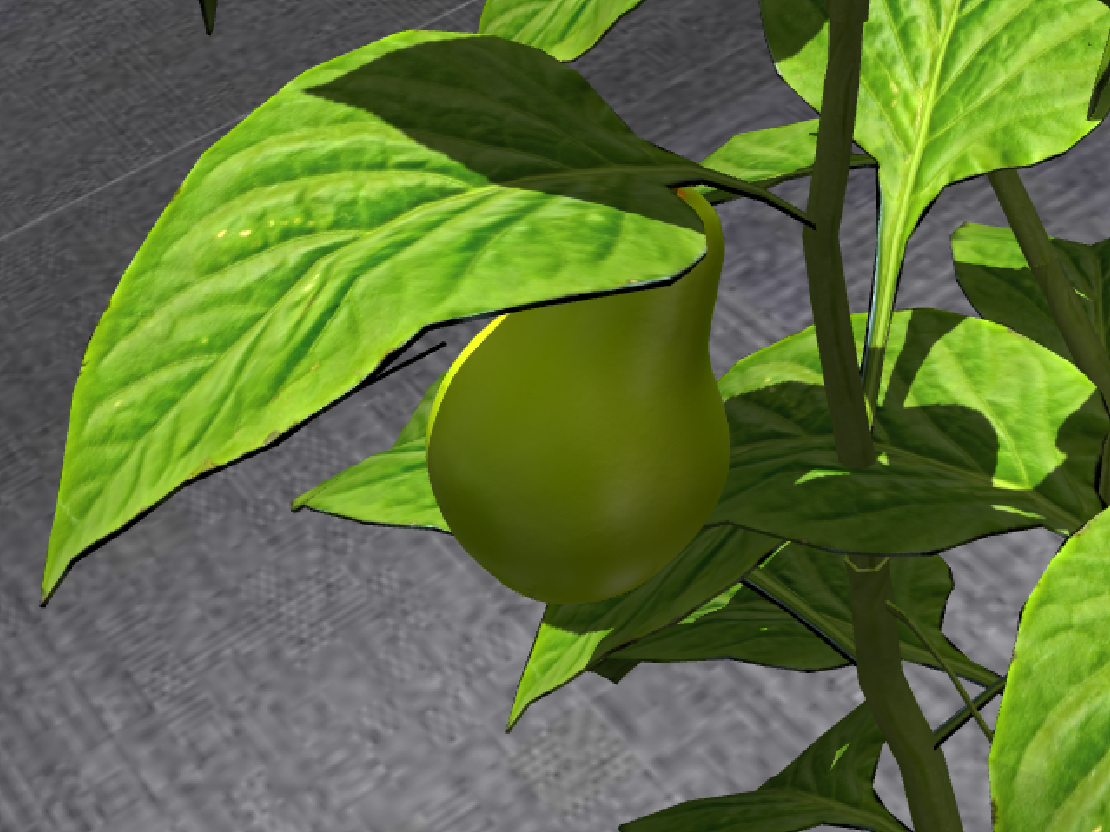}
\includegraphics[width=0.116\textwidth]{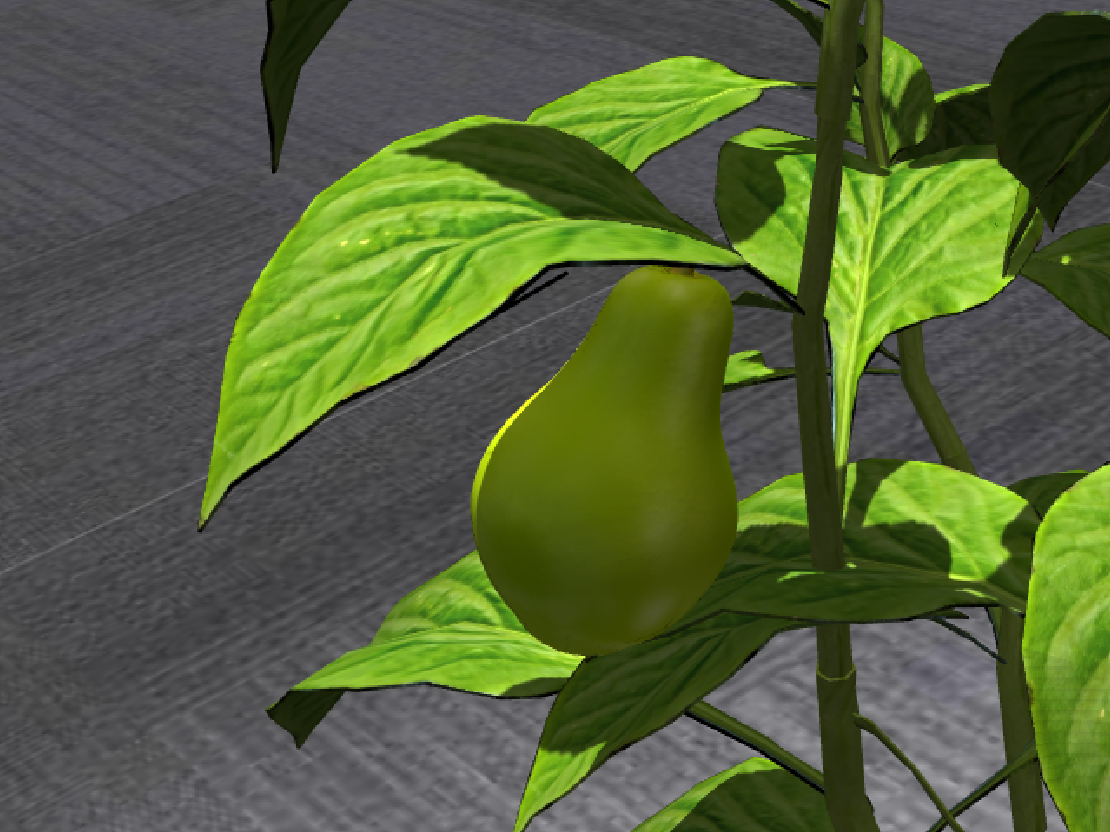}
\includegraphics[width=0.116\textwidth]{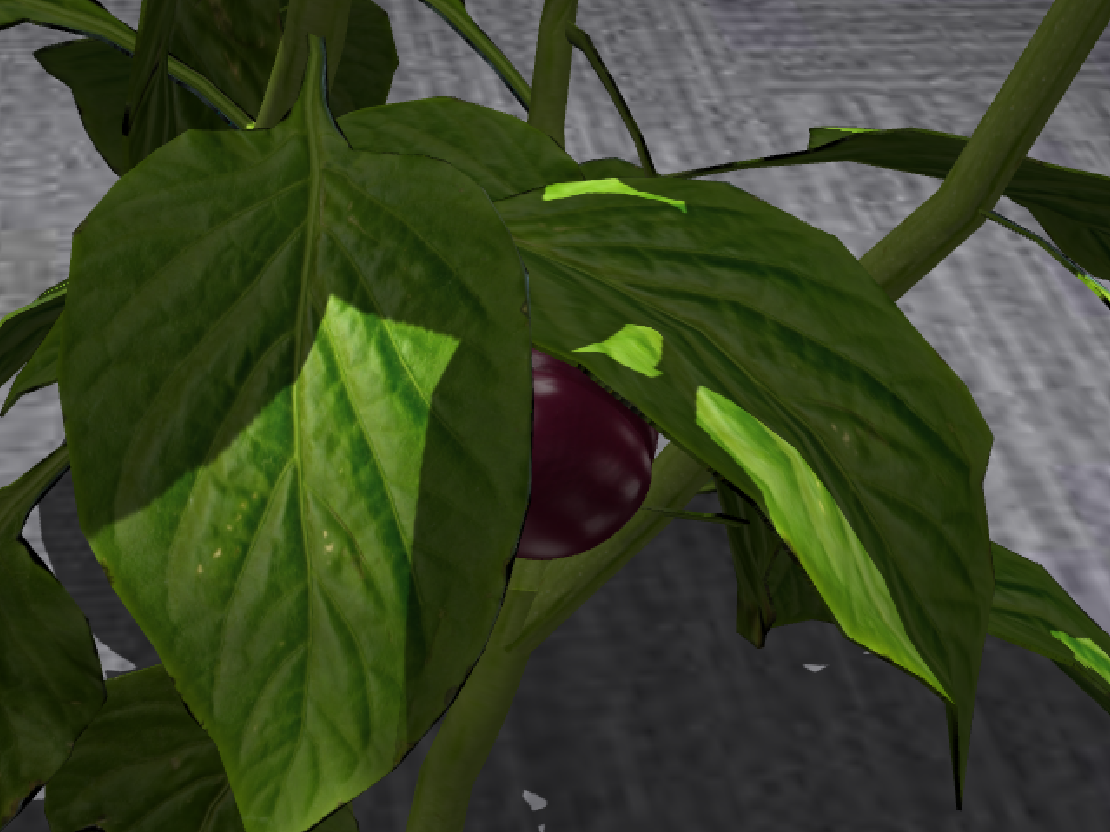}
\includegraphics[width=0.116\textwidth]{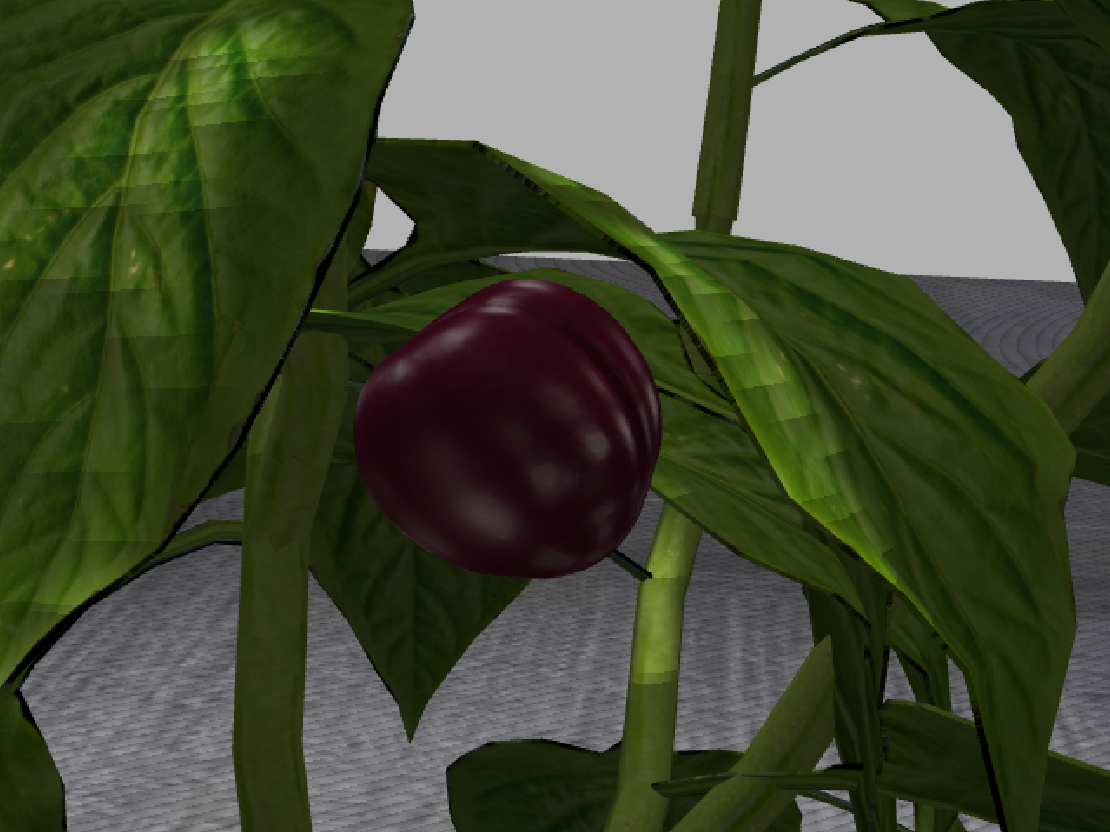}

}

\vspace{3pt}
\centerline{
\includegraphics[width=0.116\textwidth]{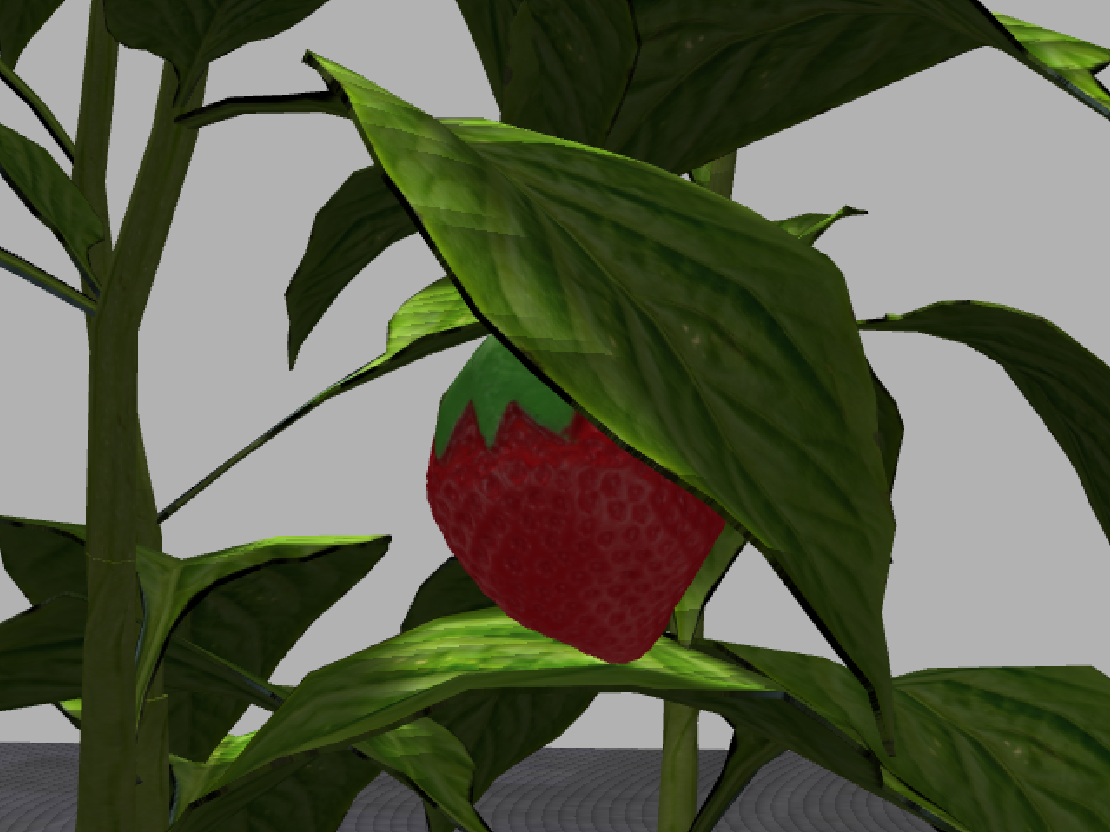}
\includegraphics[width=0.116\textwidth]{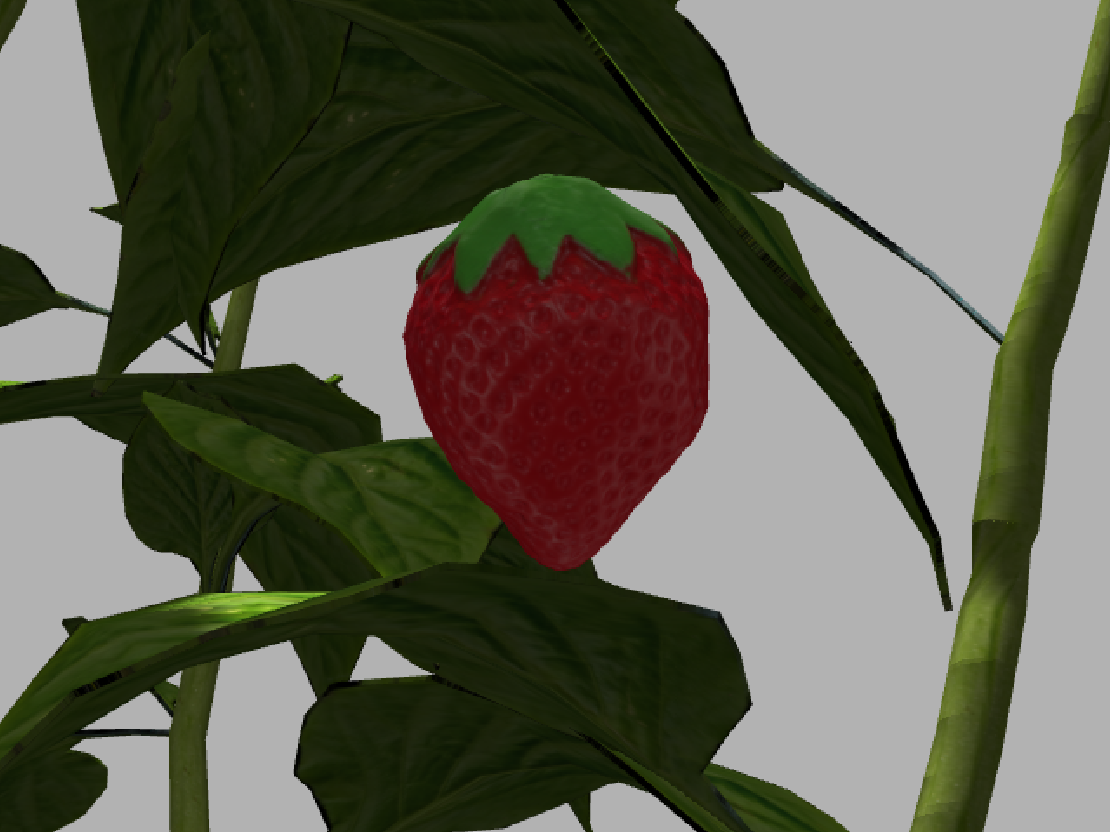}
\includegraphics[width=0.116\textwidth]{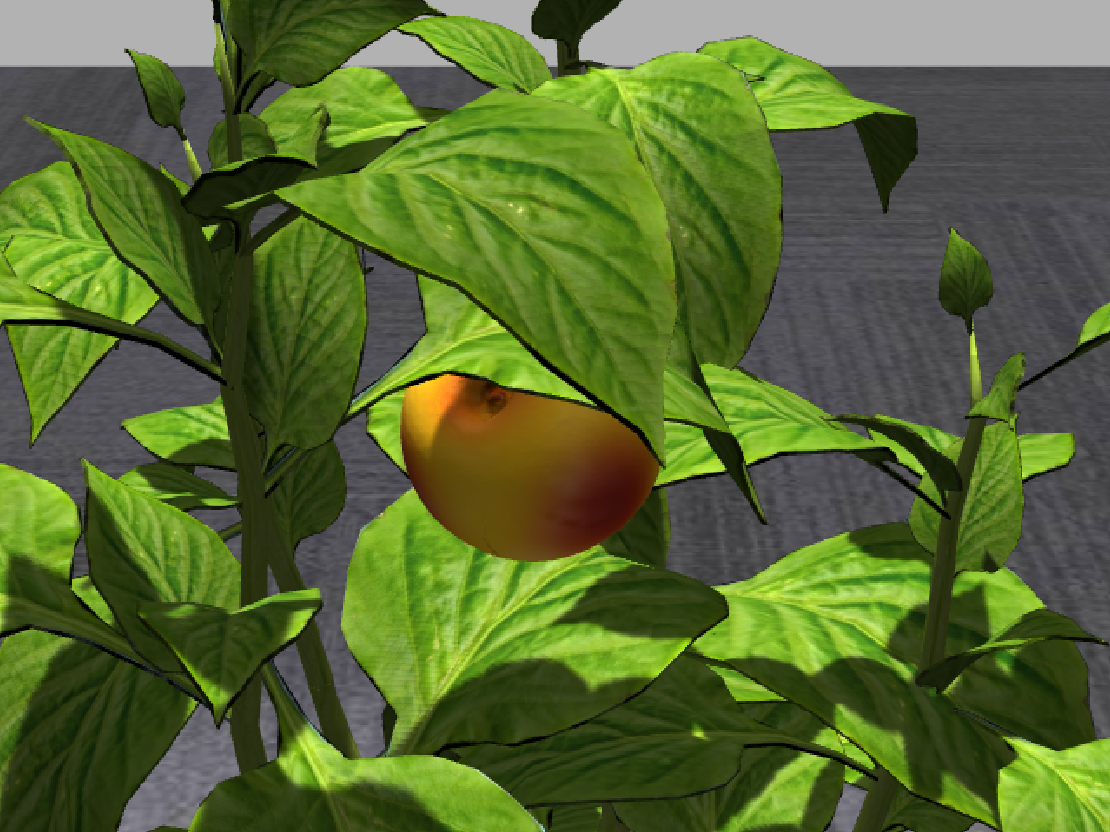}
\includegraphics[width=0.116\textwidth]{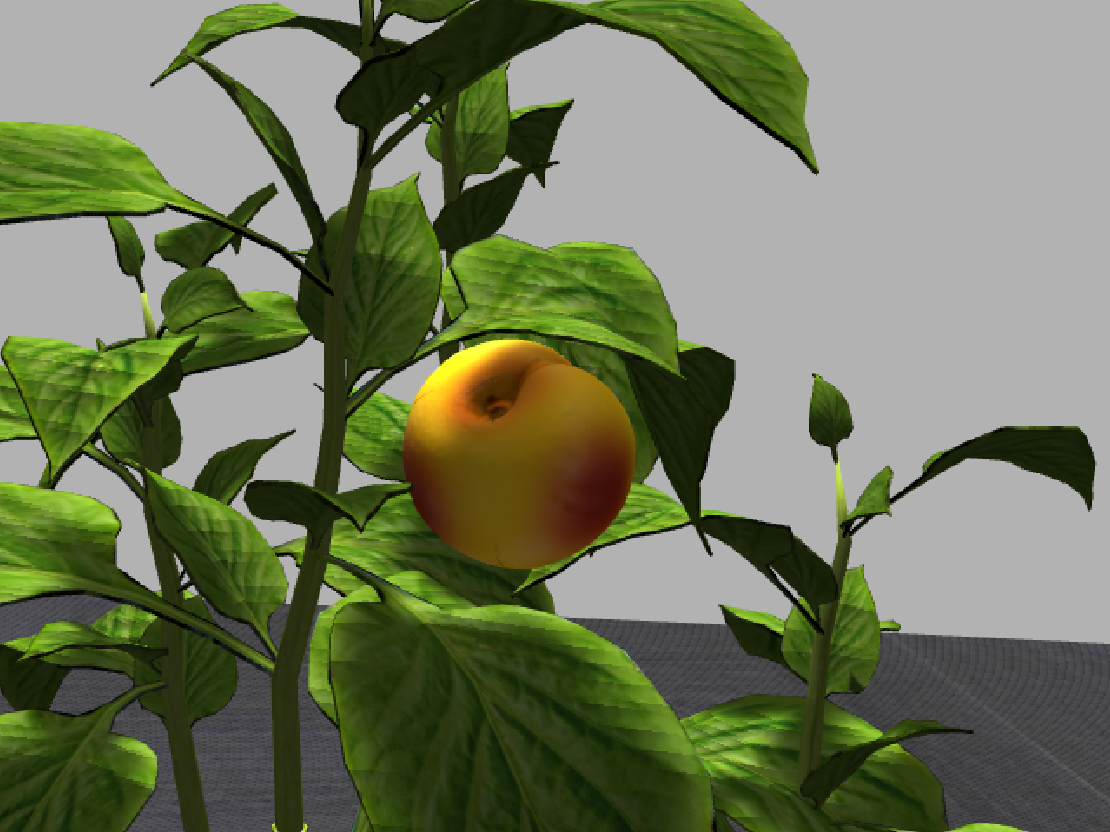}
}

\caption{
We proposed an end-to-end imitation learning-based viewpoint planning method to address the challenge of identifying ideal observation viewpoints for target crops in occluded robotic harvesting environments. Unlike existing approaches that rely on hand-engineered features, our model learns viewpoint planning policies directly from human expert demonstrations. The robot adjusts its camera pose in a continuous 6-DoF action space, enabling precise viewpoint adjustments. This leads to better visibility of occluded crops, improving fruit detection and overall harvesting efficiency across various agricultural environments.
}
\label{fig1}
\vspace{-3mm}
\end{figure}

Traditional viewpoint planning involves two main steps: view sampling and view evaluation. Various potential viewpoints are generated through probabilistic or heuristic sampling, and are then evaluated using metrics like information gain \cite{delmerico2018comparison} or task-specific objectives to determine the optimal viewpoint. However, these conventional methods have several limitations. They heavily rely on manually crafted metrics to select viewpoints, which are often tailored to specific tasks or environments, making them difficult to generalize across different scenarios. Furthermore, the quality of these metrics critically influences the performance of viewpoint planners, usually requiring tremendous expert labor for design and tuning. Most methods also assume a static environment, overlooking the dynamic nature of real-world settings, which is especially problematic in agriculture where plant growth continually alters the scene.

In recent years, machine learning-based viewpoint planning approaches have emerged in an attempt to achieve breakthroughs. For instance, Landgraf et al. \cite{landgraf2021reinforcement} applied reinforcement learning to automated inspection tasks, learning viewpoint selection policies through interaction with the environment. However, the success of reinforcement learning is highly hinged on the design of reward functions and often lacks generalization capability, mirroring the challenges of traditional feature engineering. Additionally, these methods require extensive training data and time to develop effective policies, raising efficiency concerns in practical applications. Mendoza et al. \cite{mendoza2020supervised} proposed an end-to-end deep learning framework that uses neural networks to infer the next-best-view directly from sensory data in 3D reconstruction. While effective on controlled datasets, this approach struggles to generalize to new environments, particularly in complex real-world scenarios.

Given the limitations of existing methods, we propose a novel imitation learning-based approach to address the viewpoint planning problem. Our method is designed to train robotic systems by mimicking human expert for selecting viewpoints when occlusions occur. This study aims to provide a more efficient and generalizable solution to the occlusion challenges faced in autonomous robotic harvesting, potentially overcoming the limitations of both traditional and other learning-based methods. The key contributions of this research are as follows:

\begin{itemize} 
\item A novel viewpoint planning method based on imitation learning is introduced. By emulating human expert, this method adapts to complex agricultural environments and is expected to significantly improve fruit detection and harvesting success rates, effectively addressing occlusion challenges in autonomous robotic harvesting (see Fig.~\ref{fig1}). 
\item Unlike most viewpoint planning methods that produce discrete control commands (e.g., moving up or down by 5 cm), our approach models the viewpoint planning policy in a continuous action space. This enables a smoother, more precise camera movement, which is critical for identifying optimal viewpoints in complex agricultural environments. 
\item Validation of the proposed method in both simulated and laboratory settings. Experimental results demonstrate superior performance compared to other methods, confirming the feasibility and effectiveness of our approach.
\end{itemize}

\section{Related Work}

Imitation learning, an approach that derives policies from expert demonstrations, has made significant strides in robotics and artificial intelligence in recent years. It combines the advantages of supervised and reinforcement learning, eliminating the need for complex reward function design while offering higher data efficiency, faster convergence, and greater interpretability \cite{zare2024survey}.

Behavioral Cloning (BC) \cite{pomerleau1988alvinn} is the most straightforward form of imitation learning, relying on supervised learning to map state spaces directly to action spaces by imitating expert behavior. While simple and easy to implement, BC suffers from distribution shift - discrepancies in state distributions between training and testing. To address this issue, Ross et al. \cite{ross2011reduction} introduced the DAgger (dataset aggregation) algorithm, which improves generalization through iterative data collection and policy training. More recently, Chi et al. \cite{chi2023diffusion}  made a notable breakthrough with their diffusion policy, which models visuomotor policies as conditional denoising diffusion processes and has significantly improved BC performance in complex robotic manipulation tasks. Meanwhile, the action chunking transformer (ACT) \cite{zhao2023learning}, enhances BC by segmenting action sequences into "chunks", capturing long-term behavior patterns more effectively.

Viewpoint planning is a key technology for addressing occlusion problems, with important applications in computer vision tasks such as object recognition and 3D reconstruction. Traditional viewpoint planning methods are primarily divided into search-based and synthesis-based approaches \cite{zeng2020view}. Search-based methods sample multiple candidate viewpoints and evaluate them under specific constraints, whereas synthesis-based methods directly compute the next-best-view (NBV) based on task requirements, constraints, system limitations, and sensor models. While synthesis-based methods demand fewer computational resources, search-based methods are known for their accuracy and reliability, making them the predominant approach.

While effective in structured environments, traditional viewpoint planners find it difficult to adapt to complex, dynamic settings, driving the growing popularity of learning-based approaches. Pan et al. \cite{pan2024exploiting} investigated leveraging a pre-trained 3D diffusion model as a prior for viewpoint planning from a single RGB image. Vasquez et al. \cite{vasquez2021next} developed a method using a 3D convolutional neural network (CNN) to predict NBVs from voxelized occupancy grids, eliminating the need for manual feature engineering.

In agricultural robotics, the dynamic nature of farm environments presents significant challenges. Crops vary in color, size, shape, and position, are frequently occluded by foliage, requiring careful handling. Viewpoint planning has thus become a crucial solution for mitigating occlusion. Zeng et al. \cite{zeng2022deep} used deep reinforcement learning (DRL) to determine optimal viewpoints in agricultural settings, utilizing octree-based observational maps with labeled regions of interest ROIs to encourage environment exploration. Lehnert et al. \cite{lehnert20193d} introduced 3D-Move-to-See, which employs a multi-camera array on a robotic arm to determine the optimal next viewpoint in occluded and unstructured environments. Sun et al. \cite{sun2023object} combined deep learning with active sensing to minimize occlusion in crop localization across varied scenarios. Rehman et al. \cite{rehman2021viewpoint} formulated the viewpoint planning problem as a classification problem, using a CNN-based model to guide camera positioning. In grape harvesting, Yi et al. \cite{yi2024view} introduced an active vision strategy based on spatial coverage rate, iteratively selecting the optimal viewpoint sequence for small fruit stem detection in occluded environments.

%



\section{Methodology}

This section provides a detailed explanation of our proposed imitation learning-based viewpoint planning method.

\subsection{Problem Statement and Formulation}

This study aims to develop a novel and efficient viewpoint planning approach that effectively identifies ideal viewpoints for agricultural robotic harvesting tasks, such as unobstructed observation of harvesting targets, gathering sufficient information for subsequent harvesting operations. We introduce an imitation learning-based viewpoint planner that predicts motion commands for an RGB-D camera. The camera, mounted on a robotic arm, is actuated to actively adjust viewpoints, addressing occlusion issues in agricultural scenarios. In our experiment settings, an ideal viewpoint is the one that offers a complete and unobstructed image of the target crop. Notably, the definition of an ideal viewpoint is task-specific; for instance, in strawberry harvesting, the ideal view must include the stem for successful picking. Our framework is flexible and can be adapted to specific requirements only by collecting appropriate demonstrations.

To achieve performance comparable to human experts in viewpoint planning, we employ the behavior cloning algorithm to train our policy. The entire system is modeled as a Markov Decision Process (MDP). Two types of observations are considered: a four-channel RGB-D image $i$ and a camera's six-degree-of-freedom (6-DoF) pose change $\Delta p$, representing visual and spatial perceptions respectively, denoted as $o=(i,\Delta p)$. The camera motion action $a$ has six dimensions, comprising three translational and three rotational components along the camera's local coordinate axes. This continuous action space design allows for precise and flexible camera adjustments, which are critical in complex agricultural settings.

The system operates dynamically through state transitions: at each time step $t$, given an observation $(i_{t}, \Delta p_{t})$, the robot actuates the camera to perform a motion action $a_{t}$, leading to a new state and acquiring a new observation $(i_{t+1}, \Delta p_{t+1})$. The objective is to learn a policy function $\pi$, which generates actions based on environmental observations, represented as $a_{t} = \pi(i_{t}, \Delta p_{t})$. Using behavior cloning, the policy is learned with expert demonstrations $D = \{\tau_1, \tau_2, \ldots, \tau_N\}$, where each demonstration $\tau$ is a sequence of observation-action pairs: $\tau = \{(o_0, a_0), (o_1, a_1), \ldots, (o_T, a_T)\}$. The optimal policy minimizes the difference between its predicted actions and those from expert demonstrations with respect to a specific metric:

\begin{equation}
\pi^* = \arg \min_{\pi} E_{\tau (o,a) \sim D}\left[L(\pi(o), a)\right],
\end{equation}
where $L$ is a task-specific loss function.

\subsection{Expert Demonstration Collection}

\begin{figure}[htpb]
\centerline{
\includegraphics[width=0.45\textwidth]{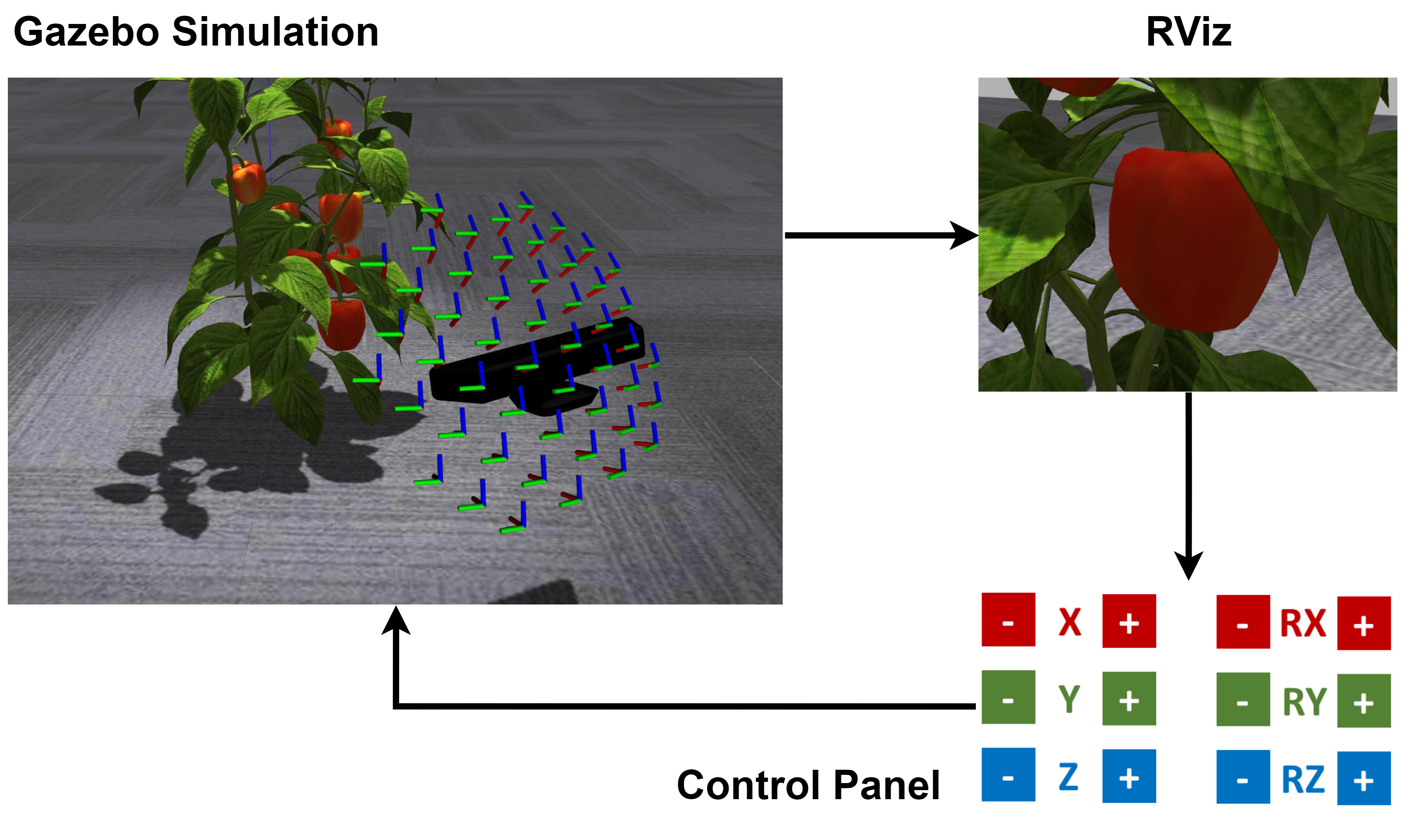}
}
\caption{We collected expert demonstration data in the Gazebo simulator as follows: For the target crop, pepper, we uniformly generated 50 initial viewpoints around it at a certain density. Starting from each initial viewpoint, an expert observed the camera's current image through the ROS RViz interface and used a custom-designed control panel to decide the corresponding 6-DoF continuous motion commands. During this process, we recorded the sequence of images captured by the camera, the camera's pose trajectory, and the smoothed continuous action commands. Together, these components form our expert demonstration dataset.}
\label{fig2}
\vspace{-3mm}
\end{figure}

To imitate human experts, collecting expert demonstration data is necessary. Taking the simulated environment built with ROS and the Gazebo simulator as an example (detailed in Section 4.1), we describe the demonstration collection process. As depicted in Fig. \ref{fig2}, for the target crop, a pepper, we assume that the harvesting robot has already approximately localized and approached the target using object detection algorithms such as YOLO \cite{redmon2016you} or Mask-RCNN \cite{he2017mask}. At this stage, a viewpoint planner is required to autonomously control camera movements for an unobstructed view of the target pepper.

Fifty initial camera poses were uniformly set in a certain density around the target pepper. From each initial viewpoint, camera images were presented to experts for data collection. Using a controller with a user interface, experts adjusted the camera’s six degrees of freedom, aiming to achieve a complete and clear view of the target crop in the image. Each demonstration consisted of up to 50 time steps, during which we recorded images, control commands, and the motion trajectory of the camera. To ensure smooth camera motion and reduce discontinuities caused by discrete time-step control, we applied smoothing to the original control commands and recalculated the necessary control commands. The resulting observation-action sequences were stored as demonstration data for training purposes. In total, 50 demonstration samples were collected, forming a comprehensive expert dataset for viewpoint planning in pepper harvesting.


\subsection{Network Architecture}

\begin{figure*}[htpb]
\centering
\includegraphics[width=0.835\textwidth]{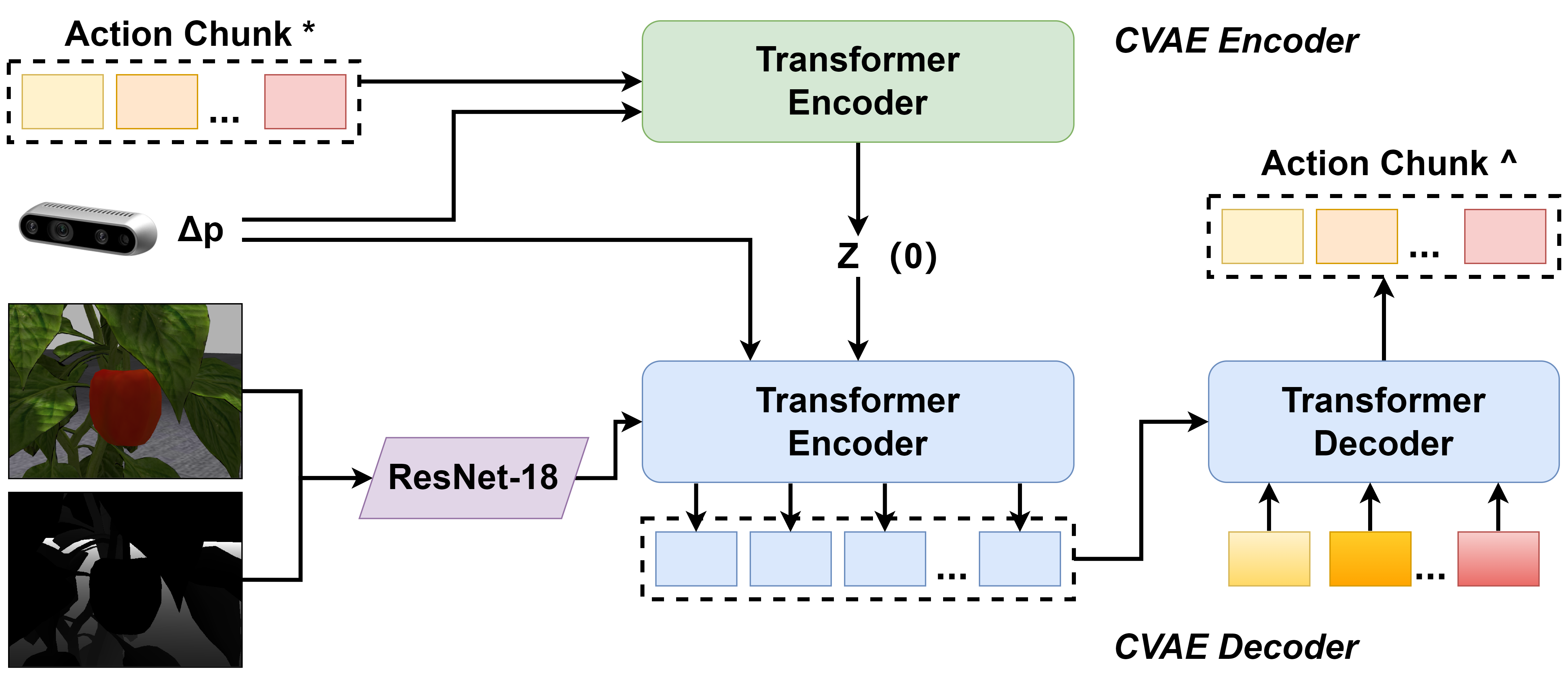}
\caption{Our network takes both RGB-D images and camera pose changes as input, and outputs refined 6-DoF camera movements. It consists of two components: a CVAE encoder and a CVAE decoder. The CVAE encoder includes a transformer encoder, which operates only during training, processing the input and generating the style variable $z$. During inference, it is discarded, and $z$ is set to a zero vector. The CVAE decoder consists of a transformer encoder and a transformer decoder, which collaboratively generate the action chunk.}
\label{fig3}
\vspace{-3mm}
\end{figure*}

We adopt the Action Chunking with Transformer (ACT) method \cite{zhao2023learning} to implement our behavior cloning algorithm. ACT addresses the compounding error problem that is common in traditional behavior cloning approaches for extended tasks. In complex robotic manipulation tasks, ACT has demonstrated superior performance over previous imitation learning algorithms. By predicting sequences of actions (action chunks) rather than individual actions, ACT reduces the effective decision-making time span. Consequently, our model evolves from a stepwise policy, $\pi(a_{t}|o_{t})$, to one that predicts action sequences, $\pi(a_{t}:a_{t+k}|o_{t})$. The final output action $a_t$ is derived using a temporal ensemble technique.

The ACT network comprises two main components: a conditional variational autoencoder (CVAE) encoder and a CVAE decoder \cite{sohn2015learning}, both leveraging the transformer architecture \cite{vaswani2017attention}. This design enables ACT to efficiently process multi-modal inputs (e.g., images and spatial positions), capture temporal dependencies in action sequences, and generate coherent, diverse actions. The self-attention mechanism of the transformer allows the model to capture long-term dependencies, while the CVAE framework supports learning and generating diverse action sequences, leading to more natural and varied robotic movements.

As illustrated in Fig \ref{fig3}, the CVAE encoder includes a transformer encoder that processes camera pose changes and a sequence of target control commands, outputting a style variable $z$. This low-dimensional latent representation captures the high-level features or style of the observations and action sequences. The CVAE decoder consists of a transformer encoder and a transformer decoder. Before feeding image data into the transformer encoder, features are first extracted using a ResNet-18 network. The transformer encoder then receives these image features, along with camera pose changes and the style variable $z$, and generates a comprehensive representation. This representation, combined with fixed positional embedding queries, is passed to the transformer decoder. The output from the transformer decoder is further processed by a multi-layer perceptron (MLP) for dimensionality reduction, and the CVAE decoder ultimately generates the action sequence, which the robot drives the camera to achieve.


\subsection{Network Training and Inference}

The network's workflow can be divided into two phases: training and inference. During training, with $k=5$, observations and sequences of actions $((i_{t}, \Delta p_{t}), a_{t}:a_{t+5})$ are sampled from the demonstration dataset. The CVAE encoder generates the style variable $z$ from these inputs, while the transformer encoder in the CVAE decoder processes the observations and $z$, and its transformer decoder predicts the action sequence $(\hat{a}_t : \hat{a}_{t+5})$. The loss function incorporates a reconstruction loss, measuring the discrepancy between predicted and demonstrated action sequences, and a Kullback-Leibler (KL) divergence, which quantifies the difference between the CVAE encoder's latent variable distribution and a standard normal distribution. The total loss is represented as:
\begin{equation}
L_{\text{total}} = L_{\text{reconstruction}} + \beta \times L_{\text{KL}},
\end{equation}
where $\beta$ is a weighting factor set to 10. This total loss is minimized using the Adam optimizer to update the network parameters.

In the inference phase, the CVAE encoder is discarded, and $z$ is set to a zero vector. For each step in a rollout, the two transformer components of the CVAE decoder collaborate to generate the action sequence. Since each prediction spans $k=5$ steps, multiple previously predicted actions contribute to the final action at time $t$. The final action $\hat{a_t}$ is obtained through a temporal ensemble as follows:
\begin{equation}
\hat{a_t} = \sum_{i=0}^{k-1} \left[ \frac{\exp(-m \times i)}{\sum_{j=0}^{k-1} \exp(-m \times j)} \right] \times \hat{a}_{t-i}, 
\end{equation} 
where $m$ is a decay factor set to 0.01, regulating the influence of earlier actions. The robot then realizes this composed camera motion action.

\section{Experiments}

This section outlines the experimental setups and evaluation metrics for our proposed method. Experiments were conducted in both simulated and real-world environments.

\begin{figure}[!b]
\centerline{
\includegraphics[width=0.475\textwidth]{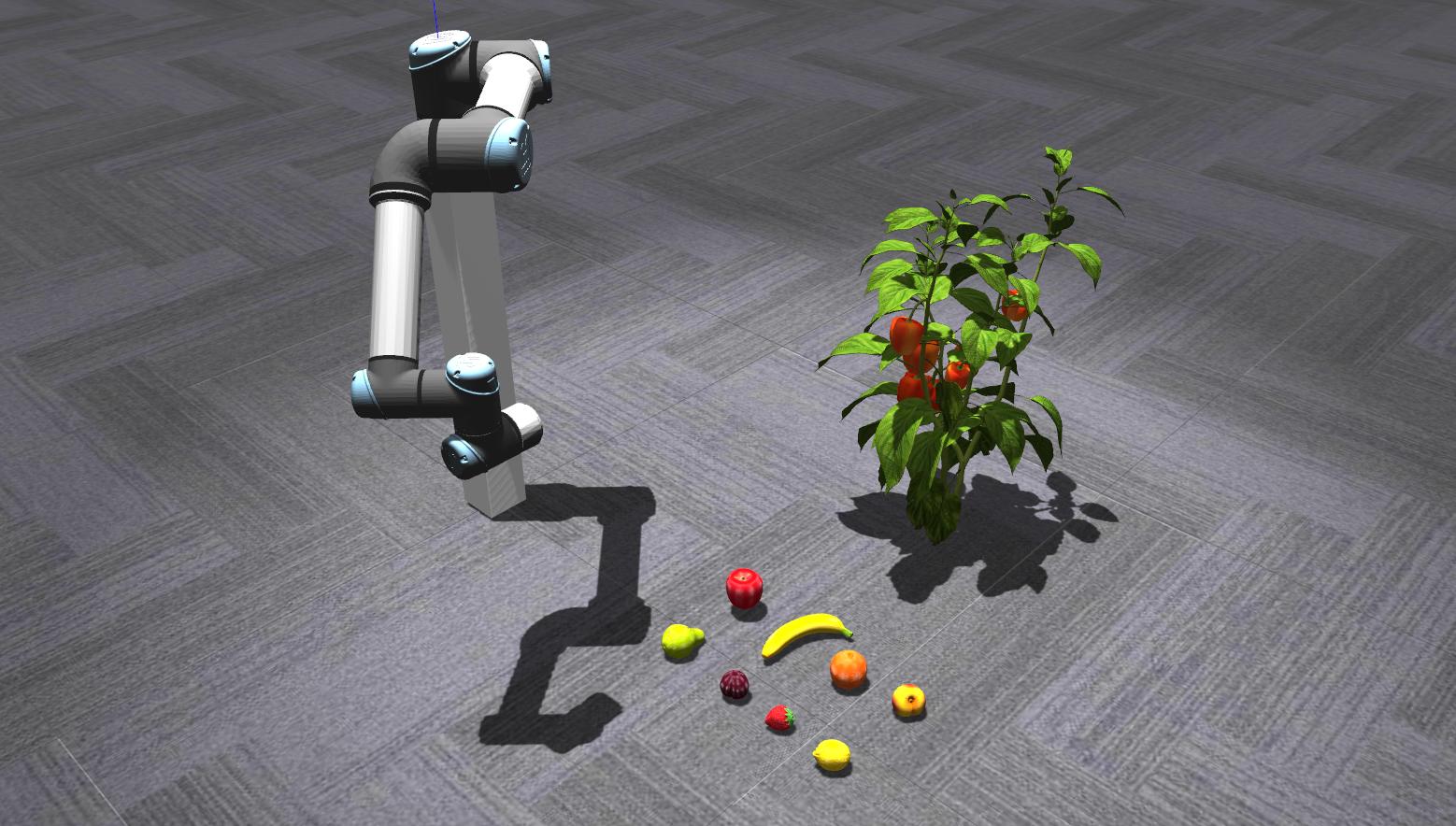}}
\caption{We built the training and experimental simulation environment using ROS and Gazebo. It incorporates a UR5e robotic arm equipped with an RGB-D camera at the end-effector, alongside a pepper plant and eight different types of fruit.}
\label{fig4}
\end{figure}

\subsection{Simulated Environments}

We constructed a simulated environment using ROS 1 and the classic Gazebo simulator to train and evaluate the proposed viewpoint planning method (see Fig.~\ref{fig4}). Building upon the work of Zaenker et al. \cite{zaenker2021viewpoint}, we extended the simulation to replicate the complex conditions found in real agricultural scenarios, particularly the occlusion of fruits by leaves.

The simulation consisted of plants bearing fruits and a robotic arm equipped with an RGB-D camera. We designed multiple scenes with varying levels of complexity, incorporating different degrees of leaf occlusion, fruit overlap, and lighting variations. In addition to peppers, we simulated occlusion scenarios for eight other fruit types to validate the generalizability of the viewpoint planner. The simulation used a 6-DoF UR5e robotic arm model, consistent with the collaborative robots employed in our real-world experiments. The robot’s end-effector was equipped with a virtual RGB-D camera to capture depth and color images of the environment.

The RGB-D image, along with the camera’s relative pose change from the previous time step, served as input to our viewpoint planner. The planner generated a sequence of 6-DoF camera motion commands $(x, y, z, roll, pitch, yaw)$ within the camera’s local coordinate system, allowing precise control over the camera’s position and orientation. These commands were then translated into joint movements for the robotic arm via the MoveIt interface, ultimately enabling accurate viewpoint adjustments.

\subsection{Experimental Setup and Results}

\begin{table}[htbp]
\centering
\caption{Comparison of Viewpoint Planning Methods}
\begin{tabular}{lcc}
\hline
\textbf{Metrics}           & \textbf{RVP}      & \textbf{Ours}      \\ \hline
Control Type               & Open-loop         & Closed-loop        \\
Control Frequency          & None              & 10Hz               \\
Average Time (s)           & 53.7              & 3.1                \\
Success Rate (\%)          & -               & 86.7             \\ \hline
\end{tabular}
\label{table:comparison}
\vspace{-1mm}
\end{table}

To evaluate the performance of the proposed viewpoint planner, we conducted experiments measuring planning success rate and execution duration time. We selected the ROI-based viewpoint planner (RVP) \cite{zaenker2021viewpoint} from agricultural context as a comparative baseline. RVP is an open-source, traditional search-based method that determines the next viewpoint based on expected information gain to construct an octree map of fruit regions.

We generated 15 occluded scenes that did not appear in the training data, using a different viewpoint generation density from that used in expert demonstration data collection. Each scene started from an initial camera viewpoint, and both algorithms were tested for up to 50 planning steps to record the number of successful attempts at finding an ideal viewpoint, with success defined as capturing the target fruit fully and unobstructed in the image. The execution time was also recorded.

The experimental results, shown in the table \ref{table:comparison}, indicate that our planner successfully identified the ideal viewpoint in 13 out of 15 tests, achieving a high success rate of 86.7\%, based on only 50 expert demonstrations. This success rate could be potentially improved further with an expanded dataset. In comparison, RVP almost always found an ideal viewpoint but did so without considering robotic workspace constraints, requiring exaggerated arm movements that are impractical for real-world applications. In contrast, our method achieves the same goal through small and refined spatial pose adjustments, similar to human behavior. 

Regarding execution time, RVP, as a discrete open-loop controller, averaged 53.7 seconds, indicating significant time consumption. By comparison, our continuous closed-loop planner operated at 10Hz, finding the ideal viewpoint in just 3.1 seconds on average in successful cases. This efficiency stems from eliminating the need to maintain an octree, reducing computational overhead, and optimizing robotic arm movements, thereby minimizing execution time.

To assess the generalization capability of our viewpoint planning method across similar tasks, we conducted expert demonstrations, training, and validation on eight different fruit types. This process was highly efficient due to minimal programming effort. In each test, the success rate and runtime were comparable, indicating that our learning-based framework is not limited to peppers but is applicable to other crops as well. 

\begin{figure}[htpb]
\centerline{
\includegraphics[width=0.24\textwidth]{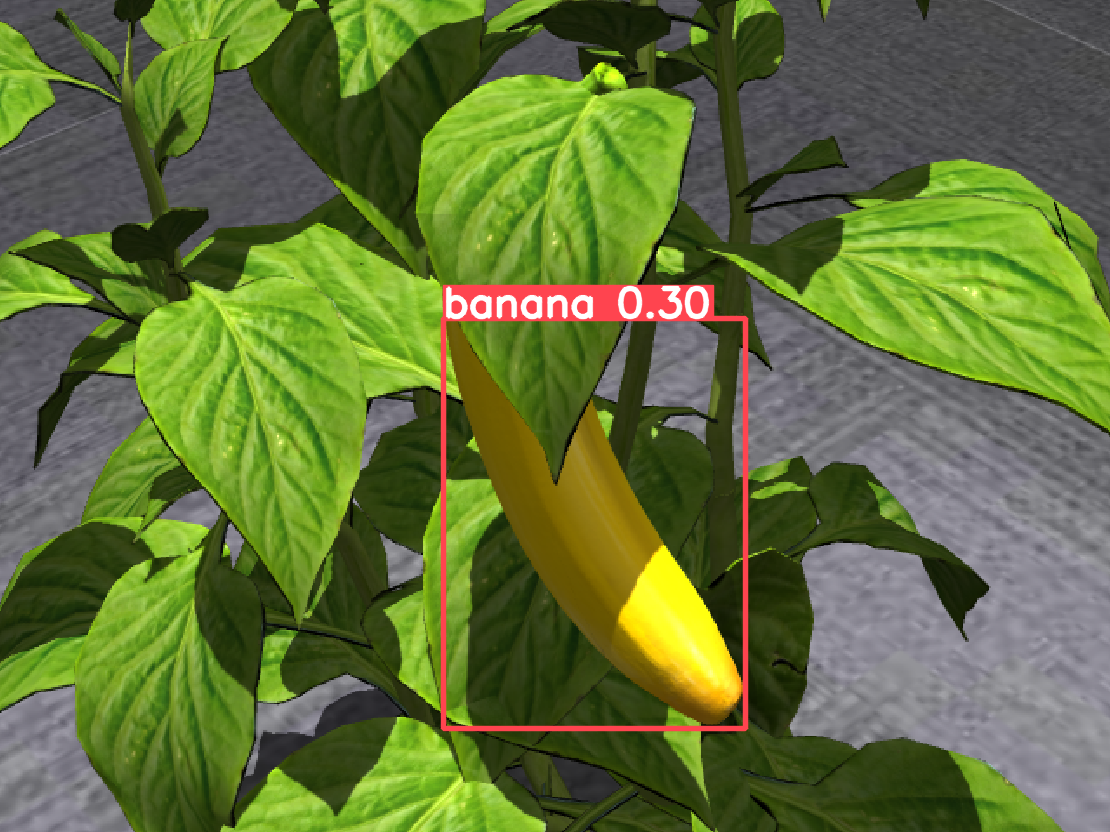}
\includegraphics[width=0.24\textwidth]{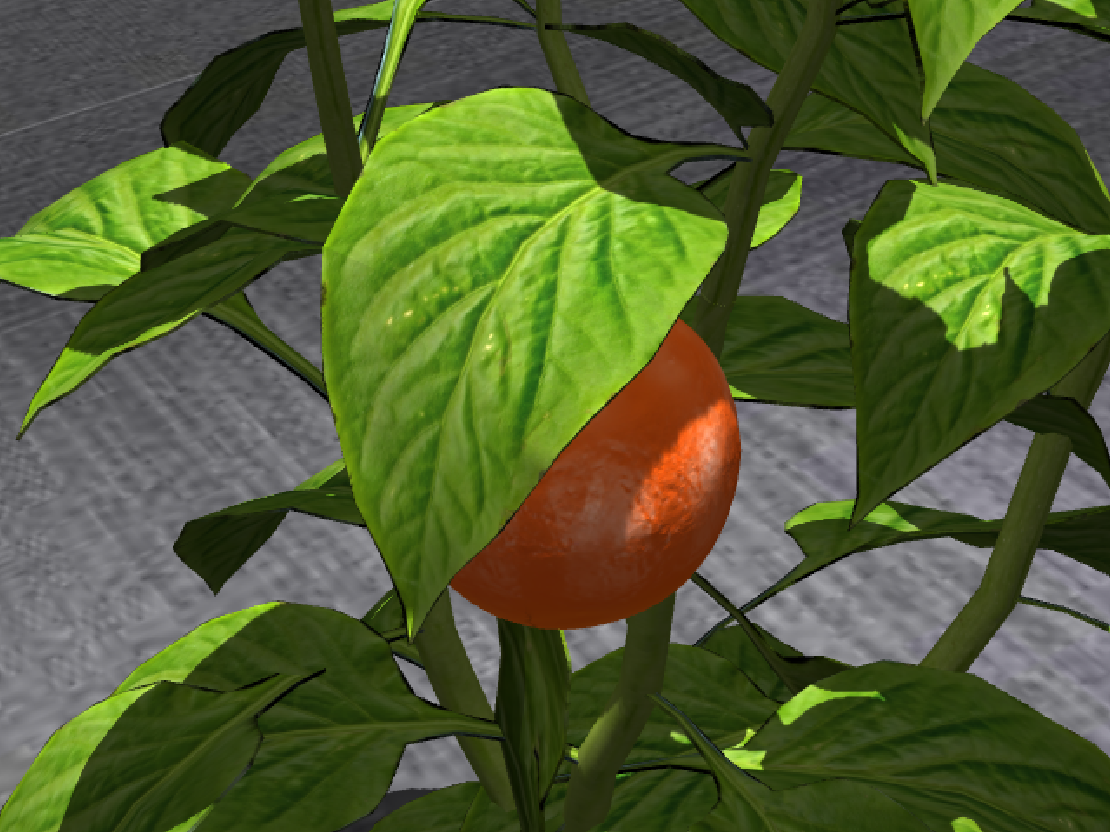}
}
\vspace{3pt}
\centerline{
\includegraphics[width=0.24\textwidth]{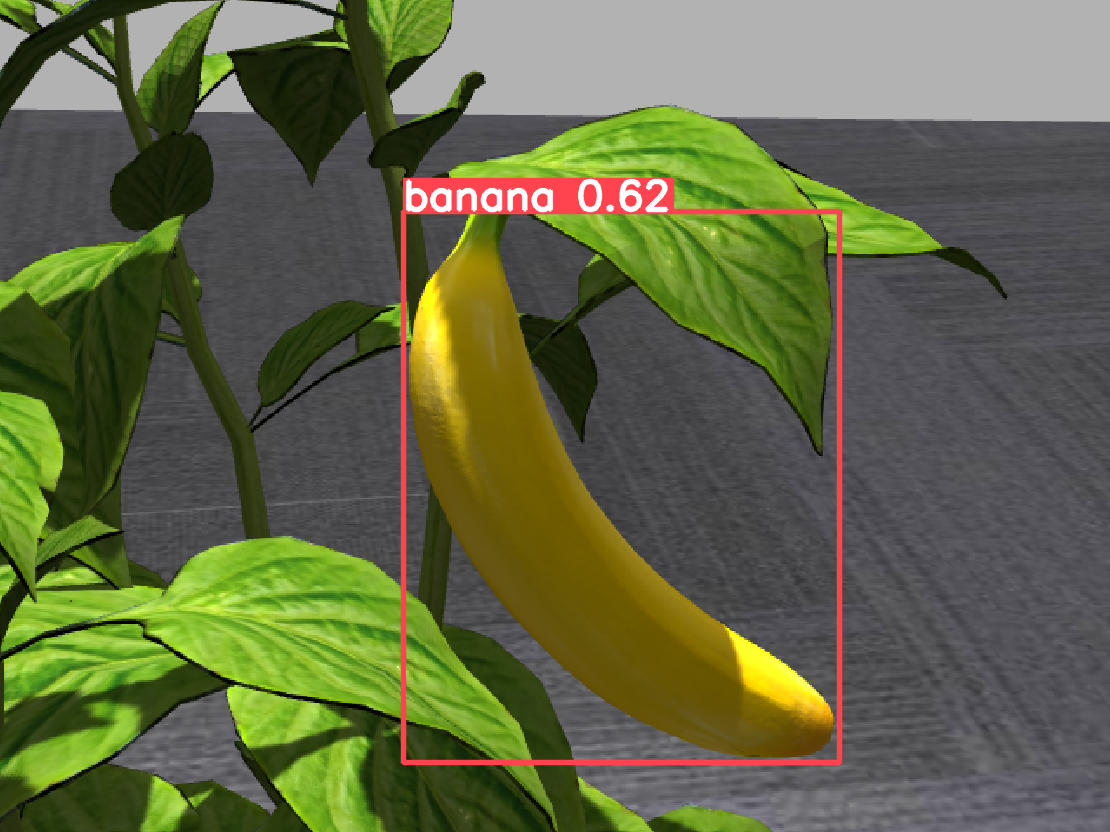}
\includegraphics[width=0.24\textwidth]{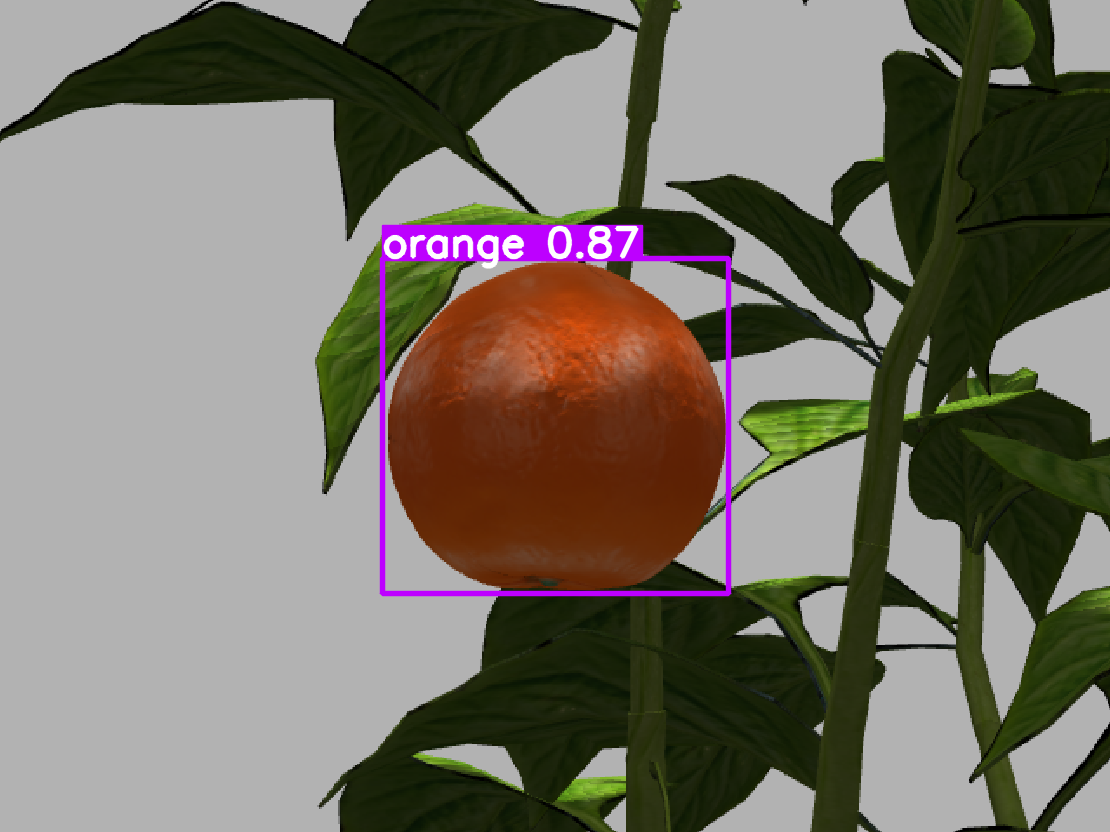}
}
\caption{Our model significantly improved bananas detection confidence from 0.30 in occluded conditions to 0.62 after adjusting to a non-occluded viewpoint. For oranges, which were undetected in the occluded scenario, the confidence score increased to 0.87 following viewpoint adjustment.}
\label{fig5}
\vspace{-3mm}
\end{figure}

Lastly, we integrated the YOLOv8 \cite{Jocher_Ultralytics_YOLO_2023} object detector to show that our viewpoint planner significantly improves detection success rates. As demonstrated in Fig.~\ref{fig5}, YOLOv8 was unable to detect the fruit from the initial viewpoint, but after the viewpoint planner adjusted the camera position, the fruit could be detected with high confidence.

\subsection{Real-world Scenarios}

\begin{figure}[!t]
\centerline{
\includegraphics[width=0.24\textwidth]{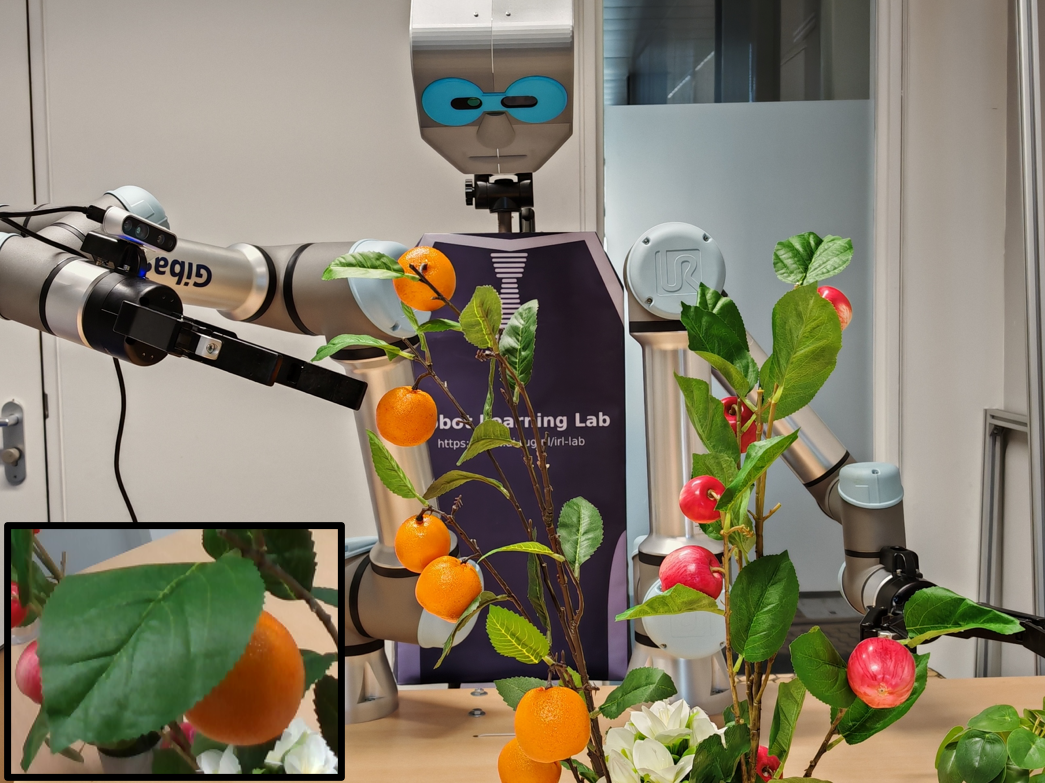}
\includegraphics[width=0.24\textwidth]{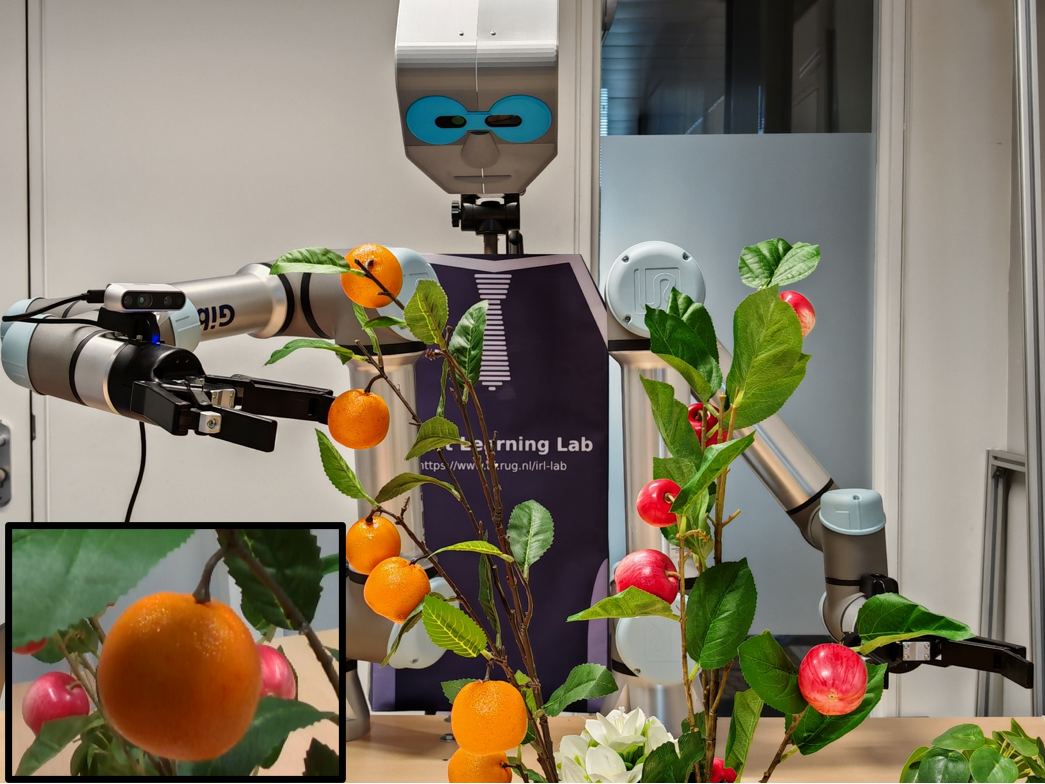}
}

\caption{Real-world experimental setup for viewpoint planning: 
(\textit{left}) The initial view shows the target fruit partially occluded by a leaf, limiting visibility; (\textit{right})
The proposed system allows the robot to perform viewpoint adjustments to tackle occlusions in agricultural environments.}
\label{fig6}
\vspace{-3mm}
\end{figure}

To validate the performance of our proposed viewpoint planning method in real-world applications, we conducted practical tests in a laboratory environment as shown in Fig.~\ref{fig6}. The experimental platform consisted of a 6-DoF collaborative robotic arm equipped with a two-finger gripper at its end effector and an Intel RealSense D435 depth camera was mounted on the robot arm to capture RGB-D images of the environment. This setup allows the robot to flexibly adjust the camera's viewpoint, simulating the process by which a human would move to search for an ideal observation position in a complex environment. Our imitation learning-based viewpoint planning algorithm was executed on a PC with an AMD Ryzen 9 5900X (12-core, 3.7 GHz) CPU and 32 GB of RAM. The experimental scenarios were designed to replicate common occlusion conditions in real agricultural environments, such as varying levels of leaf occlusion, fruit overlap, and other complexities. Multiple experiments were conducted to assess the viewpoint planner's performance across different degrees of occlusion and scene complexity.

In the real-world experiments, we collected 50 expert demonstrations for training, similar to data collection in the simulation case. The planner was then tested in 15 trials, successfully identifying the ideal viewpoint in 10 instances, resulting in a success rate of approximately 66.7\%. A key factor contributing to this success, despite the limited amount of data, was the system’s ability to make real-time adjustments to the camera’s viewpoint based on the continuous image stream, dynamically responding to occlusion levels. Moreover, the robot arm required only minimal movement to achieve these adjustments, making our method practical for real-world constraints. However, the lower success rate in real-world tests (66.7\%) compared to simulations (86.7\%) can be attributed to the more complex and cluttered backgrounds in the real environment, which made it difficult for the robot to reliably distinguish between target fruits and background objects. This occasionally led to incorrect viewpoint adjustments or prolonged search times. Additionally, variations in lighting conditions, which were not as strictly controlled as in the simulated environment, may have further contributed to some failures.

\section{Conclusion}

This study presents a novel and effective imitation learning-based approach to tackle the challenge of viewpoint planning in robotic harvesting under occlusion conditions. By leveraging expert demonstrations, our method successfully adapts to complex and dynamic agricultural environments, allowing for continuous six-degree-of-freedom camera adjustments. The results in both simulated and real-world environments demonstrate significant improvements in viewpoint planning, with our method achieving higher efficiency and accuracy compared to traditional methods. In simulated experiments, our viewpoint planner achieved an 86.7\% success rate, identifying ideal viewpoints quickly and efficiently while minimizing robot movement. The closed-loop control system, operating at 10Hz, enabled real-time adjustments, showcasing the method's capability to handle diverse occlusion scenarios. In real-world tests, although the success rate dropped to 66.7\% due to environmental complexity and clutter, the method still proved effective, demonstrating the potential for deployment in practical agricultural settings.

To further enhance the system’s performance, future research should focus on expanding the expert dataset and incorporating more diverse real-world scenarios into the training process. Interactive imitation learning could play a key role here, enabling the robot to seek human assistance during failures and progressively expand its knowledge base through corrective feedback. Additionally, exploring methods to handle background clutter and varying lighting conditions will improve robustness in real-world applications. Another promising direction is extending the framework to multi-robot systems, where multiple robots could collaborate to optimize the viewpoint planning process and increase harvesting efficiency.

\bibliographystyle{IEEEtran}
\bibliography{reference}

\end{document}